\documentclass[letter, 10 pt, conference]{ieeeconf}  

\IEEEoverridecommandlockouts                               
                                                           
\usepackage{todonotes}
\usepackage{multirow}

\usepackage{subcaption}
\usepackage{amsmath}
\usepackage{amssymb}
\usepackage{tabularray}

\let\labelindent\relax
\usepackage{enumitem}

\usepackage{stfloats}
\fnbelowfloat

\usepackage{fancyhdr}

\fancypagestyle{specialfooter}{%
  \fancyhf{}
  
  \fancyfoot[L]{\footnotesize This work was accepted by the International Conference on Systems, Man, and Cybernetics, Kuching, Malaysia, 2024.\linebreak
  © 2024 IEEE. Personal use of this material is permitted.  Permission from IEEE must be obtained for all other uses, in any current or future media, including reprinting/republishing this material for advertising or promotional purposes, creating new collective works, for resale or redistribution to servers or lists, or reuse of any copyrighted component of this work in other works.}
}

\title{\LARGE \bf
Exploring Capability-Based Control Distributions of Human-Robot Teams Through Capability Deltas: Formalization and Implications
}

\author{Nils Mandischer$^{1}$, Marcel Usai$^{2}$, Frank Flemisch$^{2}$, and Lars Mikelsons$^{1}$
\thanks{*This work was funded by the Bavarian State Ministry of Science and the Arts in the ``Augsburg AI Production Network'' as part of the High-Tech Agenda Plus.}
\thanks{$^{1}$Nils Mandischer and Lars Mikelsons are with the Chair of Mechatronics at University of Augsburg, 86159 Augsburg, Germany.
        {\tt\small nils.mandischer@uni-a.de}}%
\thanks{$^{2}$Frank Flemisch and Marcel Usai are with the Department of Balanced HSI at Fraunhofer FKIE, 53343 Wachtberg, Germany and the Institute of Industrial Engineering \& Ergonomics at RWTH Aachen University, 52062 Aachen, Germany.
        {\tt\small frank.flemisch@fkie.fraunhofer.de}}%
}

\begin{document}

\maketitle
\thispagestyle{empty}
\pagestyle{empty}

\begin{abstract}

The implicit assumption that human and autonomous agents have certain capabilities is omnipresent in modern teaming concepts. However, none formalize these capabilities in a flexible and quantifiable way. In this paper, we propose Capability Deltas, which establish a quantifiable source to craft autonomous assistance systems in which one agent takes the leader and the other the supporter role. We deduct the quantification of human capabilities based on an established assessment and documentation procedure from occupational inclusion of people with disabilities. This allows us to quantify the delta, or gap, between a team's current capability and a requirement established by a work process. The concept is then extended to the multi-dimensional capability space, which then allows to formalize compensation behavior and assess required actions by the autonomous agent.

\end{abstract}

\thispagestyle{specialfooter}

\section{Introduction}
Human-Robot Teaming (HRT) has become a new form of intuitive interaction, in which robots operate autonomously. While in classical human-robot applications, the robot had to be pre-programmed and followed a strict plan (with possible slight variations), HRT overcomes this restriction by allowing more flexibility in control authority~\cite{Varga.2024,Jadav.2024,Pacaux.2023}. In recent years, studies have been undertaken in diverse domains to study the implications of human-autonomy teaming (HAT)~\mbox{\cite{ONeill.2022,Lyons.2021}}. However, while many implicitly assume that the agents have a set of capabilities, none unify and quantify these capabilities. To give an example: In task allocation for HRTs, usually, the agents are given capabilities, or rather abstract skills, like grasping. These are neither quantifiable nor generalist, as both agents need to use different actors and the quality of a capability is usually disregarded. The unification and quantification is subject to occupational analysis systems, which are seldom used in autonomous systems. Our effort is to enable occupational analysis in HATs and HRTs by combining methodologies from diverse domains, which again enable a whole new way of interaction between human and autonomous agents. To this end, we propose a novel thought towards HRT: Capability Deltas. These are the deltas between the capabilities of a team and the process requirements stated through a work task. The capability deltas, then, may be utilized to analyze the effort needed to overcome this gap. By assigning the gap closing behavior to an automation inhibiting the supporter role, it becomes possible to assess the behavior needed to overcome this gap. This allows the automation to better anticipate human potential and to control its own behavior, or -- theoretically -- vice versa. This theory, while implemented for HAT and HRT, also holds for general human-machine interaction, as capability deltas display an essential requirement of interaction: the limitation of an agent to act.

Our main contributions are:
\begin{itemize}
    \item Formalization and quantization of capabilities and novel capability deltas, including their implications on HAT.
    \item Adaptation of established shared and cooperative control concepts towards quantitative capability analysis.
    \item Introduction of the novel capabilities-requirement (\emph{CR}) diagram for defining numeric control distributions in shared autonomy, including implications on interaction and behavior patterns.
\end{itemize}

The paper is structured as follows:
Section~\ref{sec:sota} gives an overview of predetermined motion-time systems (PMTS), introducing the \emph{IMBA}\footnote{From German ``Integration von Menschen mit Behinderung in die Arbeitswelt''.} method and capability-based HAT approaches. In Sections~\ref{sec:capabilities} and~\ref{sec:cap_deltas}, we introduce the  formalization of capabilities and their extension towards multi-dimensional capability deltas. Section~\ref{sec:exploration} shows the new formalization in a HRT exploration featuring a person with severe disabilities.

\section{Related Work}
\label{sec:sota}
The quantification methods to be discussed are based on the \emph{IMBA} method, which is used to assess the capabilities of people, particularly, with disabilities. While our methodologies are generalist, we use \emph{IMBA} to evaluate specific capabilities which are later compared to process requirements.


\subsection{Occupational Analysis Systems}
\label{ssec:occupational_analysis}
In occupational analysis systems, a work task is usually decomposed into sub-tasks. Methods-Time-Measurement (\emph{MTM})~\cite{Maynard.2012} and \emph{WORK-FACTOR}~\cite{Quick.1962} (priorly: Motion-Time-Analysis (\emph{MTA})) are two well-established standards, which define reference durations for individual actions in the sub-tasks. The individual actions are on the level of elementary human body movements, e.g., reaching with the arm. While \emph{WORK-FACTOR} was adopted into the German standard \emph{REFA}, \emph{MTM} is the de-facto standard for PMTSs in Germany. Outside of Germany, there are also other standards for PMTS, e.g, Maynard Operation Sequence Technique (MOST)~\cite{ZANDIN.2020} or the Toyota Production System (TPS)~\cite{Ono.2019}.

For matching humans to the modelled work tasks, performance assessment tools are used. \emph{O*NET}~\cite{Peterson.2001} documents a person's skills and occupational requirements, and matches both. \emph{IMBA}~\cite{IMBA.2019} adds a multitude of capabilities additional to key skill. A detailed description will be given below. Recently, Hennaert et al.~\cite{Hennaert.2022} linked \emph{IMBA} and the WHO standard International Classification of Functioning, Disability, and Health (\emph{ICF})~\cite{Ustun.2003}. Latter is a classification procedure for disabilities and illnesses. As \emph{IMBA} offers an established procedure with a comprehensive capability set, our research is based on this documentation method.

\subsection{IMBA Documentation Method}
\label{ssec:imba}
The analysis in \emph{IMBA}~\cite{IMBA.2019} is based on the quantification of human capabilities and process requirements, and their consequent comparison. The analysis guidelines are according to occupational medicine, ergonomics, and psychology. \emph{IMBA} is, particularly, aimed at the inclusion of people with disabilities, hence, establishing quantification possibilities for people with missing or lowly quantified capabilities. The broad scope of \emph{IMBA} makes it a perfect fit for the autonomous capability estimation of agents with potentially absent capabilities, e.g., in case of exhaustion, through old age, or through disabilities. In \emph{IMBA}, capabilities are clustered in nine feature complexes, e.g., see Table~\ref{tab:imba_definitions}. Each complex aggregates top-level capabilities, which again use detailed capabilities for further analysis, e.g., \emph{body posture}~(1.00) $\rightarrow$ \emph{kneeling/crouching}~(1.03) $\rightarrow$ \emph{kneeling} (1.03.01). In this work, we use a new ID system for \emph{IMBA}, where the first number is the complex, the second the top-level capability, and the third the detailed capability as displayed in the example. The third level may be ignored, if suitable. The IDs are sequenced according to the indexing in~\cite{IMBA.2019}. For the latter examples, only the complex \emph{body part movement} and two additional top-level capabilities are of importance. Table~\ref{tab:imba_definitions} lists the used capabilities and their definitions.

\subsection{Capabilities in Human-Autonomy Teaming}
\label{ssec:caps_in_hat}
Already in the 1970s, researchers covered the evaluation of human capabilities. Askren and Regulinski~\cite{Askren.1969} introduced human error metrics as means of performance reliability measures and Astrand~\cite{Astrand.1976} introduced tests to evaluate human physical capacity. Modern approaches aimed at HAT or HRT are still mostly based on testing procedures, e.g., Kolb et al.~\cite{Kolb.2021} employ two abstract cognitive tests to evaluate human performance in situational awareness and mental modelling of latent network structures in HRT scenarios. They argue that incorporating individual performance differences may improve planning approaches for heterogeneous agents. In systems with more autonomy, the testing paradigms have to shift towards sensor-based cognition. Gualtieri et al.~\cite{Gualtieri.2024} analyze cognitive ergonomics in human-robot collaboration (HRC) and develop guidelines for industrial HRC, i.a., that the automation shall understand the human skill level and capabilities, and adapt its own behavior, accordingly. They emphasize that the human shall be able to adjust the level of autonomy. LeMoyne and Mastroianni~\cite{LeMoyne.2015} quantify diverse body movement capabilities, including characteristics of Parkinson's disease using IMU data. Kahanowich and Sintov~\cite{Kahanowich.2023} learn full arm movement from IMU sensors. Mandischer et al.~\cite{Mandischer.2023} theorize to estimate capabilities based on observation using Gaussian networks and uncertainty quantification. The estimate is then used to allocate tasks between a robot and the human. However, this leads to a mutual split in task allocation. Further, only the human's capabilities are taken into consideration, disregarding the potential lack of capabilities on the team level. Recently, Chen et al.~\cite{Chen.2024} linked dynamic support roles in HRT with reaching the flow state. Within, the robot is required to assess human skills and challenges within their actions.

\begin{table}[tb]
    \caption{Selected top-level capabilities of \emph{IMBA}~\cite{IMBA.2019} used in this study. Definitions translated and IDs added. All listed capabilities inhibit more capabilities on the detailed level.}
    \centering
    \begin{tabular}{p{0.03\textwidth}|p{0.07\textwidth}p{0.07\textwidth}|p{0.23\textwidth}}
        \small\textbf{ID} & \small\textbf{Capability} & \small\textbf{Complex} & \small\textbf{Definition} \\
        \hline
        1.05 & Bent Over/ Stooped & Body Posture & Being able to adopt and maintain a posture with a bent upper body (up to 30° bent over, more than 30° stooped posture)\\
        \hline
        3.01 & Head/Neck & Body Part Movement & Being able to perform movements of the head/neck\\
        3.02 & Trunk & Body Part Movement & Being able to perform movements of the trunk\\
        3.03 & Arm & Body Part Movement & Being able to perform all movement and strength-related activities that require the use of both or one arm(s)\\
        3.04 & Hand/Finger & Body Part Movement & Being able to perform activities with the hands and fingers\\
        3.05 & Leg/Foot & Body Part Movement & Being able to perform activities with the legs and feet\\
        \hline
        9.05 & Stamina & Key Qualification & Being able to work continuously on the tasks associated with the activity without interruption
    \end{tabular}
    \label{tab:imba_definitions}
\end{table}

There is also some work linking HAT applications with the assistance of people with disabilities. Mondellini et al.~\cite{MPM23} design a HAT which supports the special behavior patterns of people with autism spectrum disorder. Kildal et al.~\cite{KMI19} support people with cognitive disabilities by highlighting parts needed in a work step through an automated light system. Thomas et al.~\cite{Thomas.2024} study the potential for human-AI tutoring in special education and find potential in supporting learning disabilities. There is virtually no work on adaptive HATs for people with disabilities focusing on physical disabilities. Most studies either focus on cognitive and learning disabilities using adaptive teams in form of human-AI and speech dialog systems, or apply rather static human-machine interaction strategies in physical work, e.g.,~\cite{HWL21,Arboleda.2020,GHF13}.

\section{Capabilities}
\label{sec:capabilities}
For the formal definition of capabilities, an evaluation system is required, which includes the requirements set for the agent's capabilities. Within, both quantified sets (capabilities and requirements) need to be assessed independently but with a common framework.

\subsection{Formalization of Human Capabilities}
\label{ssec:cap_form}
In our framework, a human agent has a set of capabilities $c_{j}$ with \mbox{$1\le j\le n$}. They are quantified for a specific human agent $i$ with the quantification range $\mathbf{Q}\subset\mathbb{N}_{0}$. The quantified capabilities are noted in the capability profile \mbox{$\mathbf{p}^{i}=(c_{j}^{i})_{j\in \mathbf{P}}$}, where \mbox{$\mathbf{P}=\{1,...,n\}\subset\mathbb{N}$} is the linear index set of $\mathbf{p}$, \mbox{$c_{j}^{i}\in\mathbf{Q}$}, and $i$ denotes a specific person or agent. Note that all human agents are evaluated on the same set of capabilities, hence, the missing index on $\mathbf{P}$. The task is defined in form of a sequence of actions \mbox{$\mathbf{b}^{k}=(r_{j}^{k})_{j\in \mathbf{B}^{k}}$}, with the requirements \mbox{$r_{j}^{k}\in\mathbf{Q}$} and where $\mathbf{B}^{k}\subseteq \mathbf{P}$. Note that the specific requirements used may change according to the action, denoted by index $k$.

\subsection{The Human Agent's Desire to Act Requires Resources}
\label{ssec:act_resources}
In a given task, the human agent is required to act in a way that uses their capabilities to a certain extent. This leads to the assumption that the human has a set of capabilities according to Section~\ref{ssec:cap_form} that they can freely dispose of. Further, we shall assume that the human has a desire to act, i.e. they want to use their capabilities in a sufficient way to fulfill the task. As already discussed by Flemisch et al.~\cite{Flemisch.2012}, such acting of capabilities requires resources. Assume a human driver trying to avoid a slowing car. Given sufficient time they may induce an evasive maneuver~\cite{Usai.2024}. However, if time is insufficient, they are majorly impaired in their (high-level) ability to control the car, which may result in an accident. Hence, even though the human desires to act, they are not able to do so. The capabilities themselves are dependent on resources, which are required to manifest a capability in the real world, i.e. we define the human to not possess a capability if they are missing the resources to act. This is in line with the works of, e.g., Whittle~\cite{Whittle.2010} or Clarke~\cite{RandolphC.2009} -- both representatives of dispositionalism. We group resources in five categories:
\begin{description}
    \item[Capability Resources] Other capabilities including their quantification.
    \item[Actuation Resources] All motor functions and actuator entities, e.g., drives or extremities, including the absence of malicious conditions\footnote{Note: Actuation Resources features only binary states, i.e. absence or presence, (continuous) gradations of actuation entities is depicted in the quantification of capabilities.}, e.g., arthritis.
    \item[Mental Resources] The mental capacity to (continuously) act a capability, analogous to \emph{stamina} (9.05) in \emph{IMBA}~\cite{IMBA.2019} (see Table~\ref{tab:imba_definitions}), or the willingness to use an opportunity to act, see, e.g., Blumberg and Pringle~\cite{Blumberg.1982b}.
    \item[Environmental Resources] Resources provided by the spatial and temporal work environment, e.g., movement space or item positions.
    \item[Societal Resources] Legal and social factors, which influence the way of work, e.g., peer pressure to refrain from using aids or \emph{OSHA} guidelines, including the regulatory influences of the work task, e.g., precision\footnote{Note: Precision requirements also require a certain concentration to act, hence, a precision requirement influences multiple categories of resources.}.
\end{description}

While analyzing the \emph{IMBA} method~\cite{IMBA.2019}, some capabilities are explicitly marked as having an influence on other capabilities. In the complex \emph{body part movement}, the two detailed capabilities \emph{rotation movements while sitting} (3.02.01) and \emph{rotation movements while standing} (3.02.02) are particularly influential, i.e. the top-level capability \emph{trunk movement} (3.02) is the most influential in \emph{body part movement}. This is, as 3.02 is used to balance certain detailed capabilities in the \emph{head/neck movement} and \emph{arm movement} capability. If we assume the human body as a star shaped tree with the root in the human's trunk\footnote{This is also used in modelling of body part relations~\cite{Yang.2011,Felzenszwa.2010}.}, it is reasonable that capabilities associated with body parts closer to the root (here: the trunk) are influential on body parts closer to the leaves in the same branch. However, while \emph{IMBA} implicates that influence is a one way relation, this is obviously not true given the example. In fact, the influence bidirectional: the neck can compensate trunk movement limitations and the trunk can compensate neck movement limitations. Ultimately, influence is established by pairs of capabilities, which we will later term conjugated capabilities. The categories of other resources (actuation, mental, environmental, societal) are directly derived from the influential factors in the \emph{IMBA} capabilities. Note that in contrast to the capability resources, the capabilities usually do not influence these resources, e.g., an arm movement will not influence the availability of the arm itself.
\begin{figure}[b!]
    \centering
    \includegraphics[width=.43\textwidth]{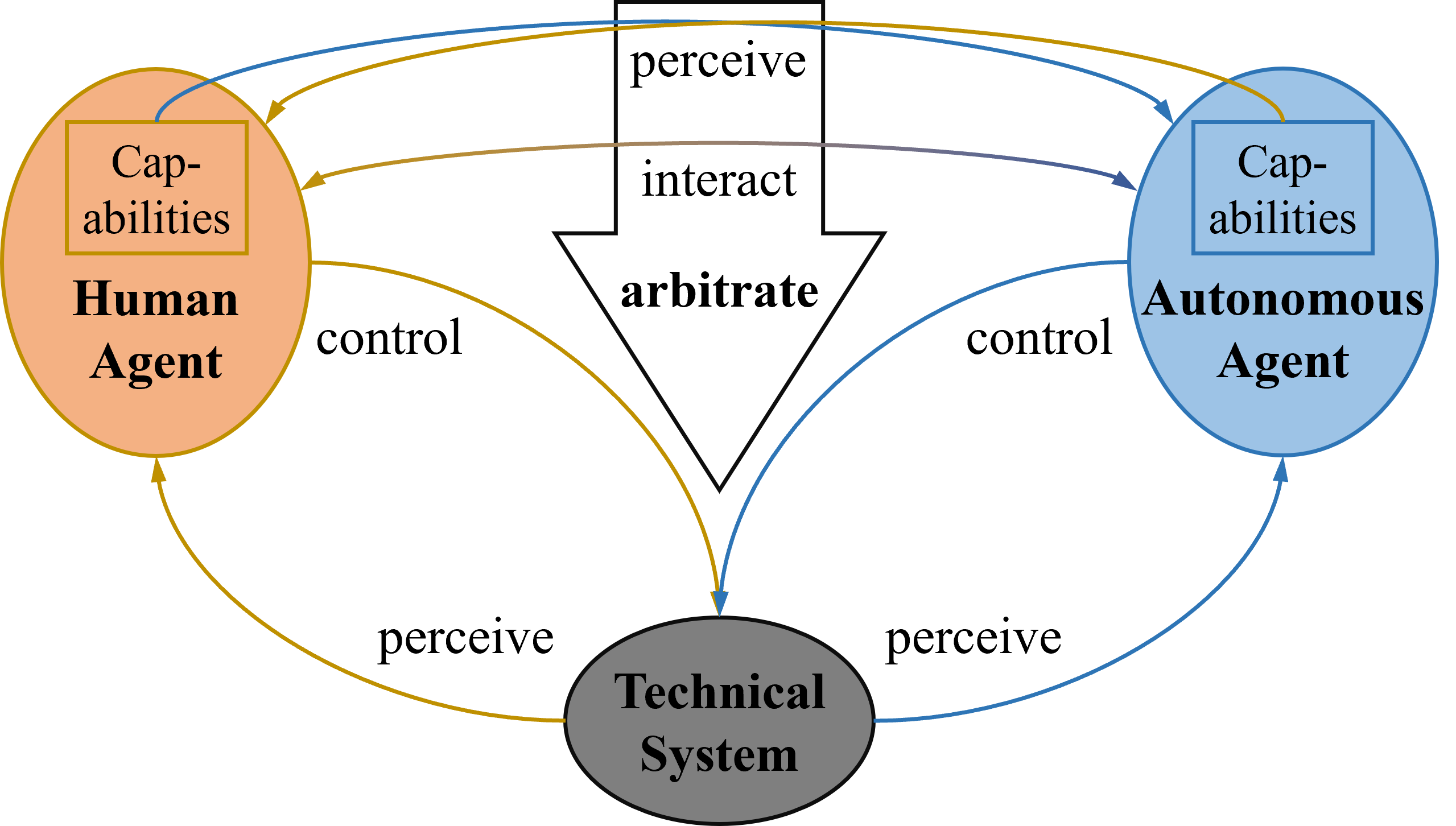}
    \caption{Arbitration of a shared task. Flemisch et al.~\cite{Flemisch.2012} extended by the agents perceiving their individual capabilities. Arrows indicate information flow.}
    \label{fig:rhoy_extended}
\end{figure}
\begin{figure*}[t!]
    \centering
    \includegraphics[width=.87\textwidth]{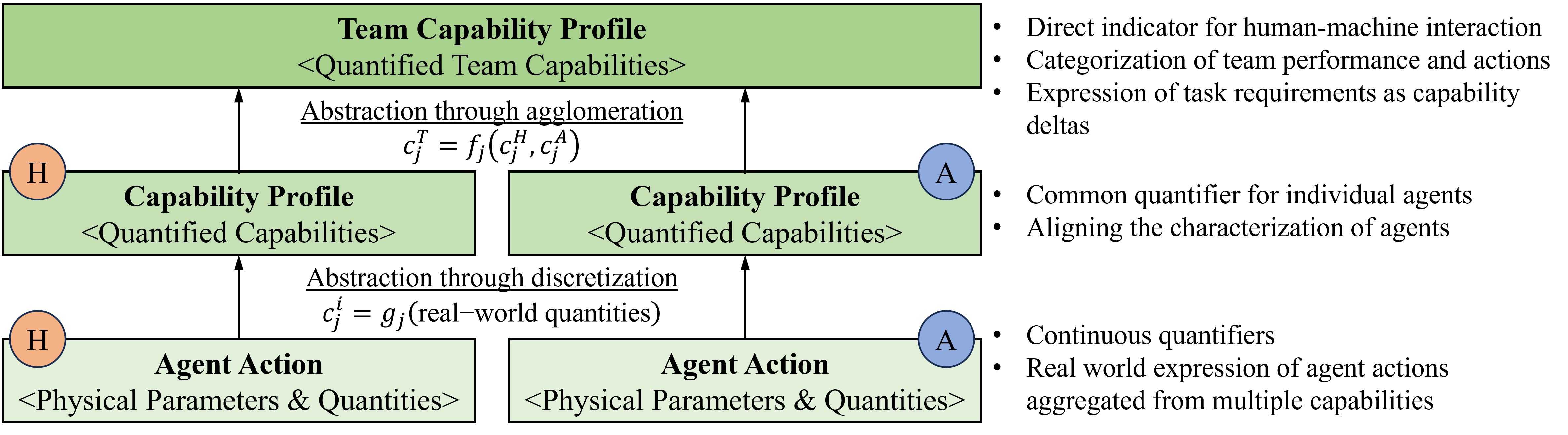}
    \caption{Layers of capability abstraction from real-world quantities to team capabilities.}
    \label{fig:layers}
\end{figure*}

\subsection{Relation of Human and Autonomous Agent: Perceive, Interact, Arbitrate}
\label{ssec:arbitration}
Until now, we have only considered the human agent. However, we can model the autonomous agent analogously. But how do both agents find a consensus on a control distribution? Arbitration, which was used prominently in previous work~\cite{Flemisch.2020, Flemisch.2012}, describes the process of negotiating the agents' own control desire through interaction to reach a consensus from an initial dissonance. Thus, the key to arbitration is that agents interact with each other to align their intent. However, it was disregarded, on which base and knowledge this interaction takes place. We extend the previous pattern to incorporate the agents' capabilities (see Figure~\ref{fig:rhoy_extended}). Each agent perceives not only the technical system and its environment, but also the capabilities of the other agent. This takes place by means of observing the other agent's behavior and individual decisions in the shared task. The perceived capabilities, then, state the base to arbitrate a shared control decision.

Without an understanding of the other agent's capabilities, it is impossible to craft good control actions and derive own decisions in the arbitration. This behavior is also observed in everyday work life. It is the essential job of every team leader to assess their team members' capabilities and use them to their optimal potential. The same is possible in a HAT: through arbitration while perceiving the other agent's capabilities. Therefore, cognition of capabilities can be understood as a base to later shared task models and shared decision making. Note, that there are capabilities which are not observable within the arbitration process. Understanding of these, however, is beyond the scope of this paper.

\section{Capability Deltas}
\label{sec:cap_deltas}
While Section~\ref{sec:capabilities} introduced the formalization of individual capabilities and their rationale on how action desire may be perceived in task negotiation, to control a human-autonomy or human-robot system, the task requirements have to be taken into consideration. If a quantified capability of a human is insufficient for a work task, the team has to cooperate to still fulfill the given task. This requires a shared understanding of what lacks in task fulfillment, which is formalized in the Capability Deltas. Note that the following description is centered about $1 \times 1$ human-autonomy teams. However, the same methods may be extended to $n\times m$ teams of an arbitrary number of human and autonomous agents or to homogeneous interaction of teams of only human or autonomous agents. Note that we assume that one group of agents are leaders and the others support the leaders when they lack the capabilities to fulfill the task, i.e. capability deltas are quantified on the team level but agents need to close the gap by means of their individual capabilities. Hence, the situation usually occurs that one agent lacks the capability (leader) and the other supports (support) in order to bring back the interaction system in a controllable state, in which the team capability is sufficient. Roles in teaming were comprehensively covered by other researchers, e.g., human teams~\cite{Stefanov.2009} or human-robot teams~\cite{Chen.2024}.

\subsection{Formalization of Capability Deltas}
A capability delta is the gap between a team's combined capability and the process requirement. The team's capability $c_{j}^{T}$ is expressed as
\begin{equation}
    c_{j}^{T} = f_{j}(c_{j}^{H},c_{j}^{A}),
    \label{eq:team_cap}
\end{equation}
where \mbox{$c_{j}^{H}\in\mathbf{p}^{H}$} and \mbox{$c_{j}^{A}\in\mathbf{p}^{A}$} are the capability of the human and the autonomous agent, respectively. Both are combined by means of any control distribution function $f_{j}(.)$ for capability $j$ defined on $(\mathbb{N}_{0})^{2}$. Note that while $c_{j}^{i}$ denotes the \textbf{capacity} -- or the maximal value an agent may act (without context) --, usually they act less in a given task and within context, hence, we commonly use the \textbf{performance} \mbox{$0\le\hat{c}_{j}^{i}\le c_{j}^{i}$} instead. Further note, there is no such differentiation for the team's capability, as it cannot be isolated from the teaming context. The wording of capacity and performance is inspired by the \emph{ICF} standard, referring to an agent without and within context, respectively.

A capability delta is defined by the scalar difference
\begin{equation}
    \delta_{j}^{Tk} = r_{j}^{k} - c_{j}^{T}
    \label{eq:cap_delta_scalar}
\end{equation}
between a team's quantified capability and a specific requirement, where \mbox{$\delta_{j}^{Tk}\in\{-\max(\mathbf{Q}),...,\max(\mathbf{Q})\}\subset\mathbb{Z}$}, \mbox{$j\in \mathbf{B}^{k}$}, and \mbox{$r_{j}^{k}\in\mathbf{b}^{k}$}. Consequently, a capability delta is the linear margin a team's capability needs to be raised to fulfill a given requirement (\mbox{$\delta_{j}^{Tk}>0$}) or by which the team exceeds the requirement (\mbox{$\delta_{j}^{Tk}<0$}). Note, that by defining $\delta_{j}^{Tk}$ on the team level, it's scale is independent of the chosen control distribution function $f_{j}$ (Figure~\ref{fig:layers}). The scalar expression in Equation~\ref{eq:cap_delta_scalar} is extended to the full vector of requirements, establishing the multi-dimensional capability delta
\begin{equation}
    \mathbf{\Delta}^{Tk} = (r_{j}^{k}-c_{j}^{T})_{j\in\mathbf{B}^{k}}.
    \label{eq:full_delta}
\end{equation}
Note, that Equation~\ref{eq:full_delta} operates only on a subset of the full capability set $\mathbf{P}$. The relation described by the capability deltas expresses a highly abstracted view of real-world parameters. This relation is depicted in Figure~\ref{fig:layers}. Note that the team capabilities are only the abstraction of the agglomerated agent capabilities. Hence, \mbox{Equation~\ref{eq:team_cap}} only agglomerates the individual capabilities but has no direct implications on how the real-world quantities are abstracted. Instead, there must exist another function $g_{j}(.)$ which translates a mixture of real-world quantities into capabilities of type $j$. By this, $g_{j}(.)$ is a mapping which also models the metric of the discretization, e.g., if the scale is linear or quadratic.

\subsection{Classification of Capabilities -- Two Examples}
To underline the importance of modelling the team's capability as an arbitrary function (Equation~\ref{eq:team_cap}), we have to consider how assistance is applied to each individual team member. We differentiate summative and non-summative capabilities. Summative capabilities are aggregated by a linear function $f_{j}$, which allows the team's performance to be higher than the individuals' capacity. Non-summative capabilities are aggregated by a non-linear function $f_{j}$ and the maximal individual capacity defines the maximal team performance.

Imagine a task in which a human and an autonomous agent must cooperatively lift an object. The involved capability $lifting$ (5.01) is usually linear \mbox{\textbf{summative}}\footnote{Assuming the same direction of the applied force vectors, uniform density and volume, and neglecting the leverage effect.}, i.e. the involved forces are directly summative, hence,
\begin{equation}
    F^{T} = F^{H} + F^{A} \Leftrightarrow c_{5.01}^{T} = \hat{c}_{5.01}^{H} + \hat{c}_{5.01}^{A}.
    \label{eq:linear_summative}
\end{equation}
For this, we have to assume a linear scale for $c_{5.01}$, or
\begin{equation}
    c_{5.01} = g_{5.01}(F) = \max_{b\le |F|/a}(b),
\end{equation}
where \mbox{$a>0$} is the width of a discretized category in $c_{5.01}$ and \mbox{$b\le\max(\mathbf{Q})\in\mathbb{N}_{0}$}. A linear summative capability implies \mbox{$\hat{c}_{j}^{A}\equiv\delta_{j}^{Tk}$} while the automation takes the supporter role. 

The majority of capabilities, however, are \mbox{\textbf{non-summative}}. Imagine a task in which a human and an autonomous agent must collaboratively reach for an object. If the human fails to reach the object, the autonomous agent cannot close this gap by just applying the difference. Instead, the autonomous agent must have a capacity sufficient to reach the object itself. In the supportive action, only a fraction of the capacity is applied in form of the capability delta. Therefore, in non-summative capabilities, $c_{j}^{i}\ge r_{j}^{k}$
states an additional requirement to the control distribution, where $i$ is any involved agent (in the example, $i=A$).

\subsection{Autonomous Assistance for Individual Capabilities}
\begin{figure}[b!]
    \centering
    \includegraphics[width=.36\textwidth]{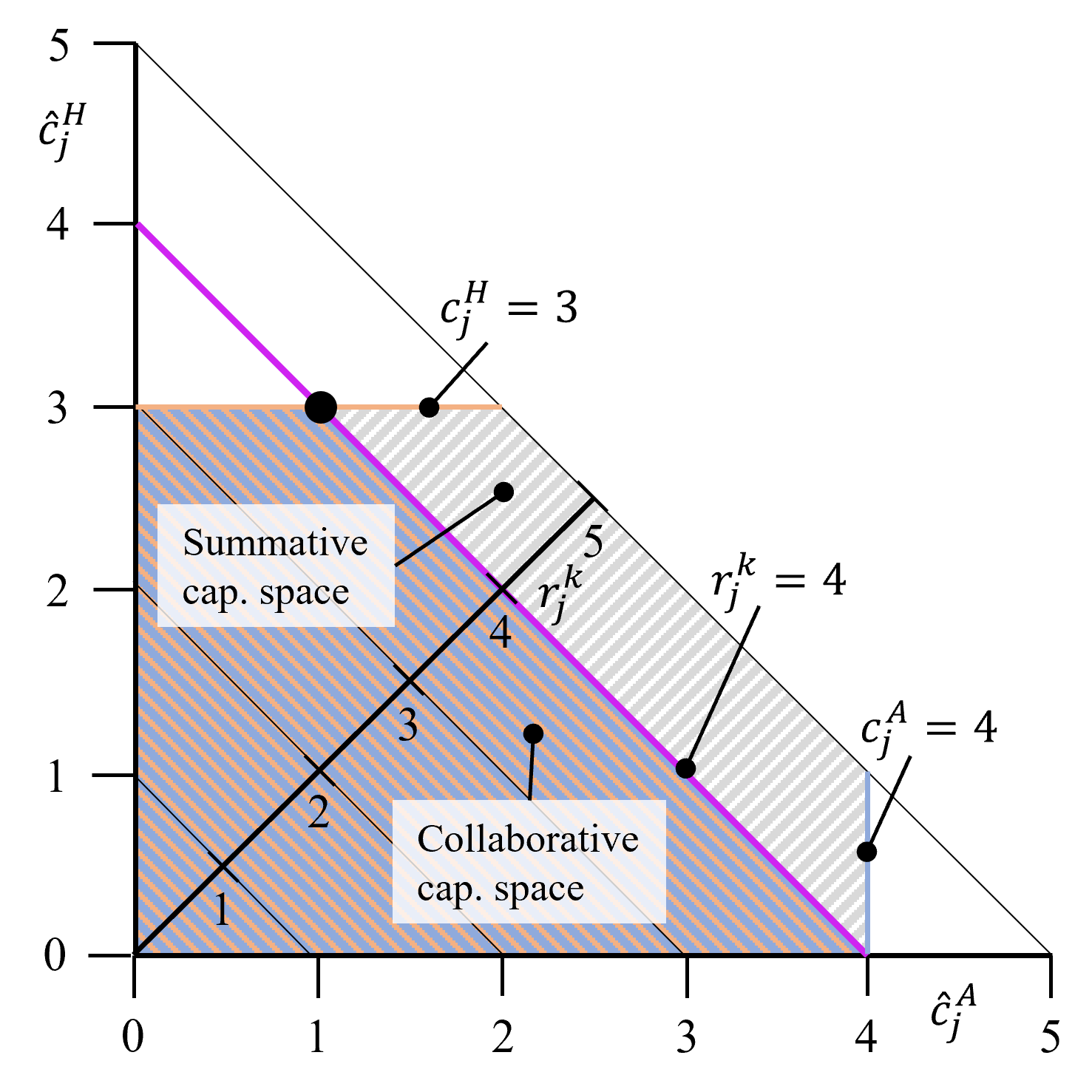}
    \caption{Exemplary \emph{CR} diagram based on the \emph{IMBA} scale (\mbox{$\max(\mathbf{Q})=5$}). A control distribution may be chosen in the collaborative capability space for any capability. Given a summative capability, also the summative capability space may be used. A feasible control distribution shall be chosen on the requirement line (here $f_{j}(c_{j}^{T})\equiv r_{j}^{k}=4$). Note that a $f_{j}$ similar to Equation~\ref{eq:linear_summative} is used. In this example, the team can fulfill the requirement given any type of capability.}
    \label{fig:yi_diagram}
\end{figure}

To find suitable control distributions over numeric capabilities, we introduce the novel capabilities-requirement, or \emph{CR}, diagram (see Figure~\ref{fig:yi_diagram}). In the \emph{CR} diagram, the human agent's capability is annotated on the vertical axis and the automation's capability on the horizontal axis. The distribution control function $f_{j}(.)$ is annotated in the diagram space as requirement $r_{j}^{k}$. In Figure~\ref{fig:yi_diagram}, a linear distribution is depicted, but theoretically an arbitrarily shaped function $f_{j}(.)$ could be depicted. The capacities of the individual agents are annotated on the individual axes. The area enclosed by the capacities depicts the potential shared control distributions. This is split into the collaborative capability space, where all type of capabilities are controllable, and the summative capability space, in which feasible control distributions may only be crafted in case of summative capabilites. A capability delta, while not explicitly depicted, is the orthogonal distance from the requirements line to the chosen control distribution (dot). In the example in Figure~\ref{fig:yi_diagram}, a feasible control distribution may always be crafted given $r_{j}^{k}=4$, hence, $\delta_{j}^{Tk}=0$. However, in some situations, it may be feasible to use parts of the \emph{CR} diagram outside the optimal control distributions, i.e. outside the requirement line.

\subsubsection{Interpretation of the CR Diagram and Capability Deltas}
\label{sssec:interpretation}
Based on the \emph{CR} diagram, we can make some observations. Firstly, it is likely that at least one agent's capacity needs to match or exceed the requirement. This is as the majority of capabilities are non-summative. Notably, the most control distributions are located outside the capacity of either agent, i.e. agents usually use a performance different from their individual capacity. In case of missing capabilities, the capability delta indicates the effort to lift the team's capability onto the requirement line. By this, it also indicates the distance from the optimal control distribution (also in case of an over-fulfillment, i.e. $\delta_{j}^{Tk}<0$). Secondly, there are potentially multiple control distributions which fulfill the requirement, including over-ful-fillment. Further, the \emph{IMBA} method implicates that some tasks are still controllable with minor degeneration in an individual capability. Therefore, it is subject to future research, if the minimal capability delta (i.e. \mbox{$||\Delta_{j}^{Tk}||\rightarrow 0$}) is always the best possible objective. Arguably, this might be the best approach, as a low capability delta keeps the human agent involved the most, promoting acceptance and motivation, or does not continuously overburden the human agent. Thirdly, even tough a control distribution may be derived, behavior patterns of the autonomous agent are missing that close the gap indicated by the capability deltas. Note that it is virtually impossible to craft a task that only involves a singular capability.

\subsubsection{Capability Deltas as Indicator of Control Deficit}
As indicated in the \emph{CR} diagram, not only the human, but also the autonomous agent may have missing capabilities. The most obvious aspect is, that static agents (e.g., collaborative robots) are missing any form of base -- or body -- movement. In \emph{IMBA}, this would be equal to zeroes in the whole complex \emph{body movement} (i.e. \emph{walking/ascending}, \emph{climbing}, \emph{crawling/sliding}). Hence, even given the aspects discussed in Section~\ref{sssec:interpretation}, there might be cases in which no feasible control distribution may be found. Flemisch et al.~\cite{Flemisch.2012} term this as ``control deficit'', i.e. both agents as a team do (at least temporarily) not have full control of the shared task. Hence, a shared action becomes impossible. While just observing individual capability deltas, there is no possibility to overcome this limitation.

\subsection{The Delta Compensation Pattern}
\label{sssec:delta_compensation}
\begin{figure}[tb]
    \centering
    \includegraphics[width=.37\textwidth]{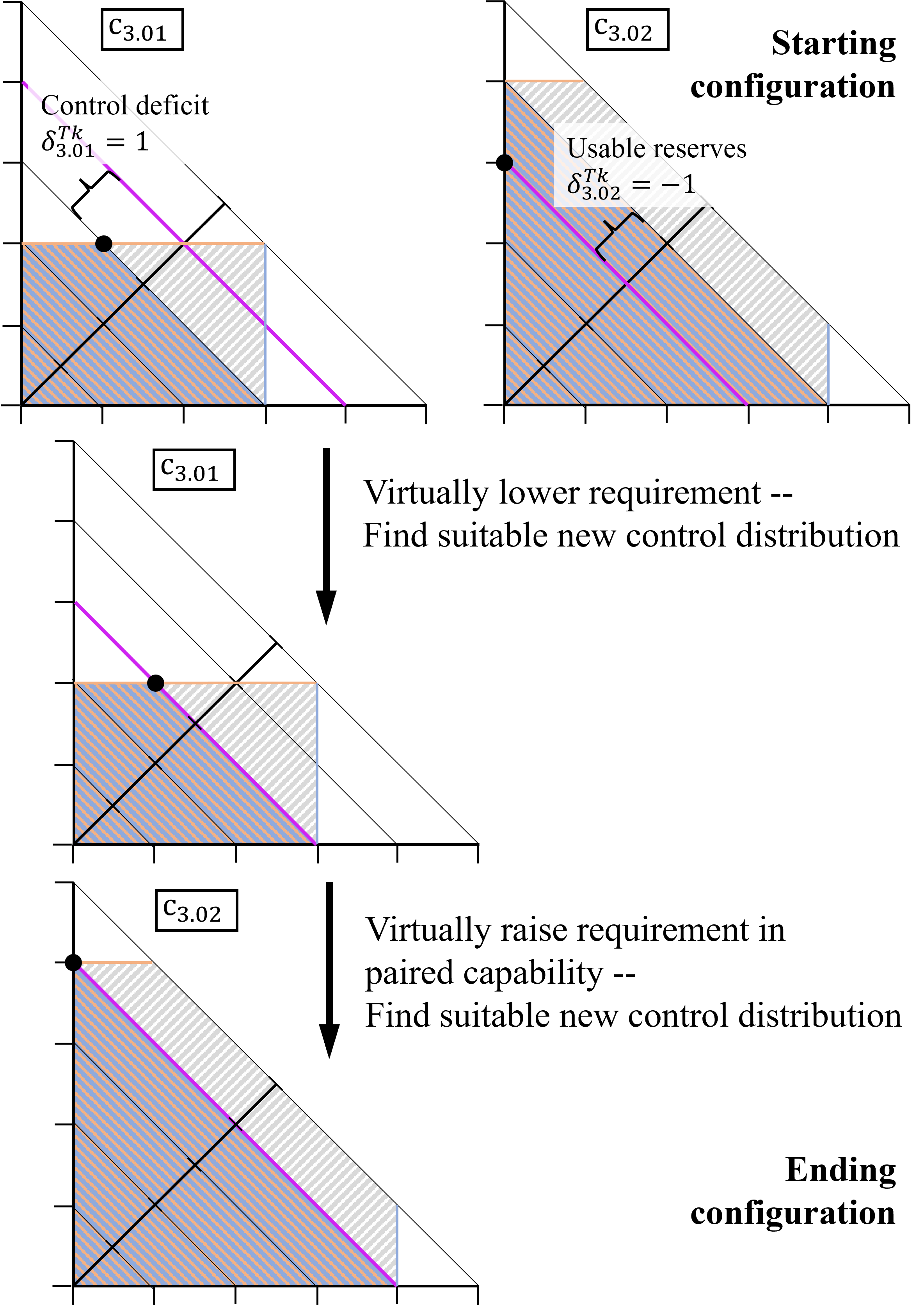}
    \caption{Delta compensation pattern for two conjugated capabilities. The black dot marks the chosen control distributions. We choose distributions with highest human contribution.}
    \label{fig:delta_compension}
\end{figure}

To solve the lack of sufficient team capabilities, multiple individual capability deltas have to be consulted. In the following, we introduce a pattern, which may be used to find suitable control distributions in the multi-dimensional capability space. In \emph{IMBA}, it is intended that missing capabilities may be compensated. Commonly, this is done by changing the framework conditions of the work system, e.g., by bringing items closer to the human. Additionally, capabilities may be used to compensate each other. In Section~\ref{ssec:act_resources}, we indicated that capabilities may influence other capabilities by means of resources. In \emph{IMBA}, i.a., the rotational movement of the trunk (3.02.01, 3.02.02) influences the head and neck rotation (3.01.01) and the backwards reach (3.03.07, 3.03.08). There are always multiple capabilities involved in solving a task. For compensating capability deltas, we can build influencing pairs of conjugated capabilities, e.g., $(c_{3.01},c_{3.02})$ or $(c_{3.02},c_{3.03})$. If any involved team capability degrades ($\delta_{j}^{Tk}>0$), the system is brought into a controllable configuration by virtually lowering the requirement in this capability (Figure~\ref{fig:delta_compension}). To compensate the now missing global task fulfillment, the requirement in a conjugated capability is raised. This requires, that the conjugated capability has usable reserves, indicated by a negative capability delta given the agents' capacities.

\subsection{Implications of Capability Deltas on Control Authority}
Capability deltas implicate, that one agent limits the team's capability and that the other agent is mainly tasked with compensating the capability delta, analogous to leader and supporter roles. Latter is the one with more usable reserves, usually associated with the autonomous agent. Therefore, it is reasonable to assume that in the majority of these kind of HAT applications, the autonomous agent will be in charge of deciding on the control distribution, i.e. the autonomous agent has the majority or all of control authority. On one hand, this contrasts the theory of Flemisch et al.~\cite{Flemisch.2012}, in which the ability to control must not be smaller than the control authority. On the other hand, the work of F\"{u}gener et al.~\cite{Fugener.2022} supports that human-AI teams perform better if the AI is in charge of changing the control authority. However, there is no indication whether this also impacts the individual responsibility of the agents, or if this may create a shortcoming between the human's control authority and responsibility. Latter aspects have been discussed in the field of Meaningful Human Control~\cite{DeSio.2018}, where such shortcoming is desired to help the society adopt machines with higher levels of autonomy. It remains to be seen if the common theory holds when autonomous agents improve their cognitive abilities and are enabled to better argue on the human's capabilities. To this end, we currently describe the kind of sketched interaction as unbalanced but refrain from taking any conclusions in this regard.
\begin{figure*}[t!]
    \def\picwidth{.8\textwidth}
    \def\toppicwidth{.8\textwidth}
    \centering
     \begin{subfigure}[t]{0.245\textwidth}
         \centering
         \includegraphics[width=\toppicwidth]{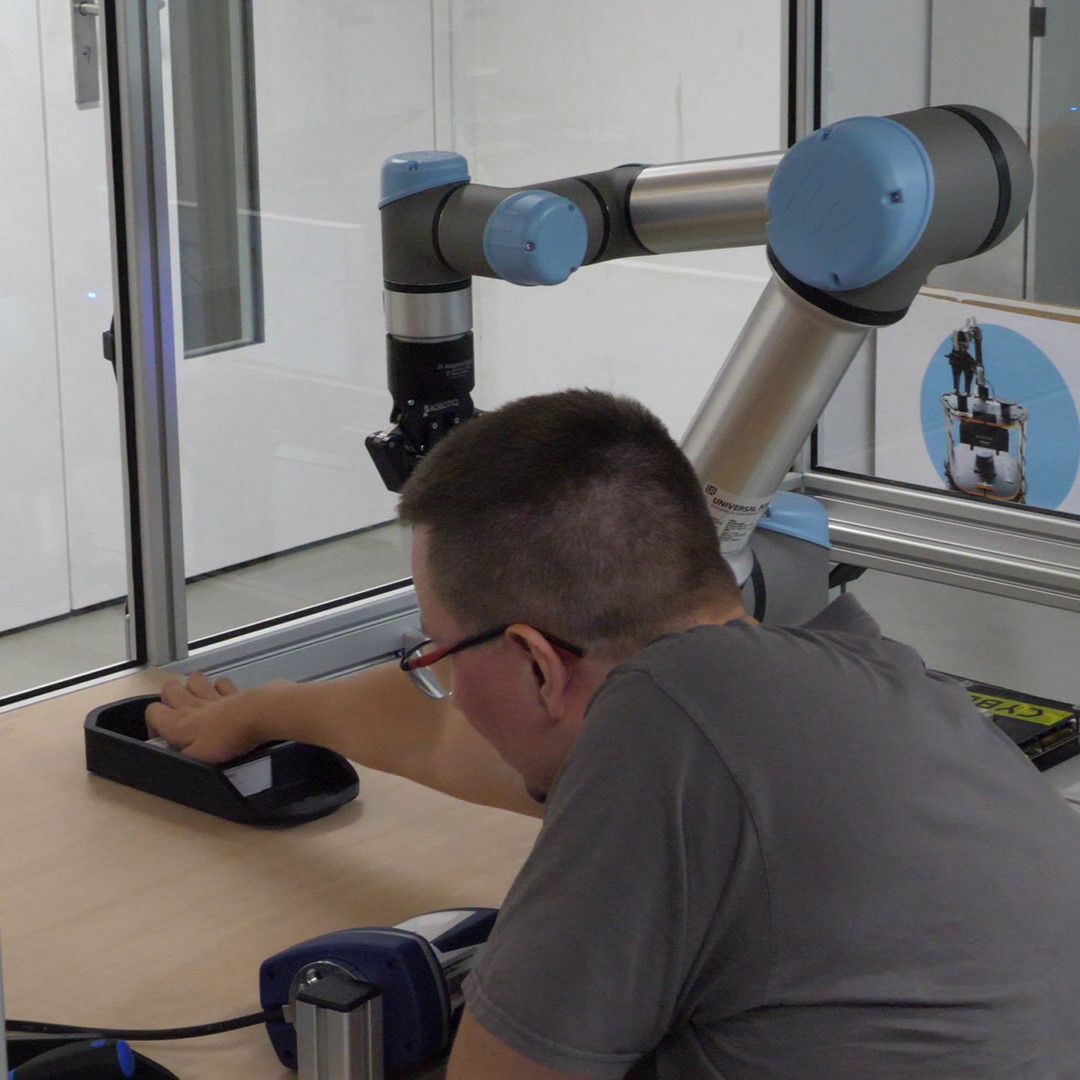}\\
         \vspace{2mm}
         \includegraphics[width=\picwidth]{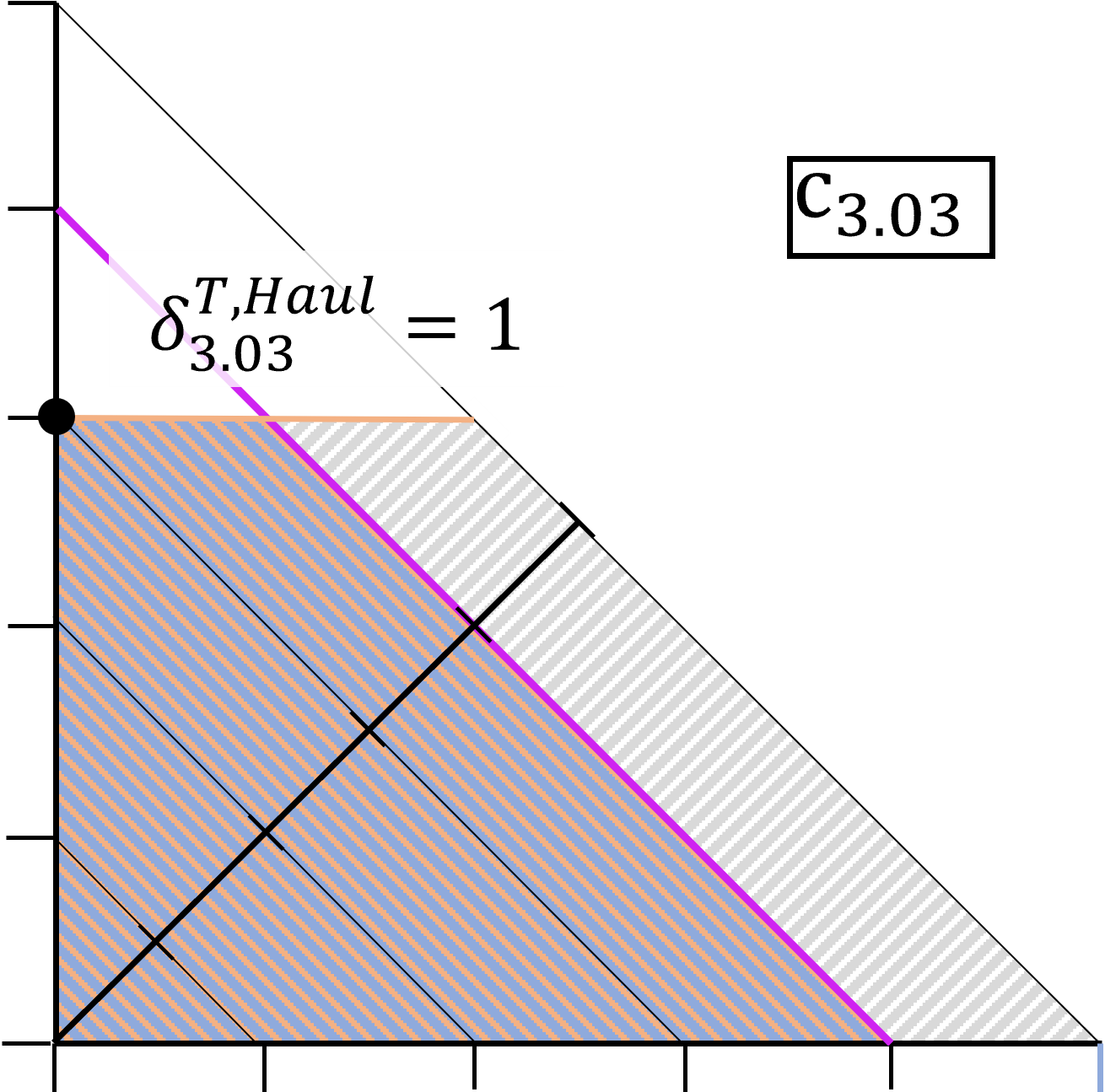}
         \label{sfig:expl_only}
     \end{subfigure}
     \hfill
     \begin{subfigure}[t]{0.245\textwidth}
         \centering
         \includegraphics[width=\toppicwidth]{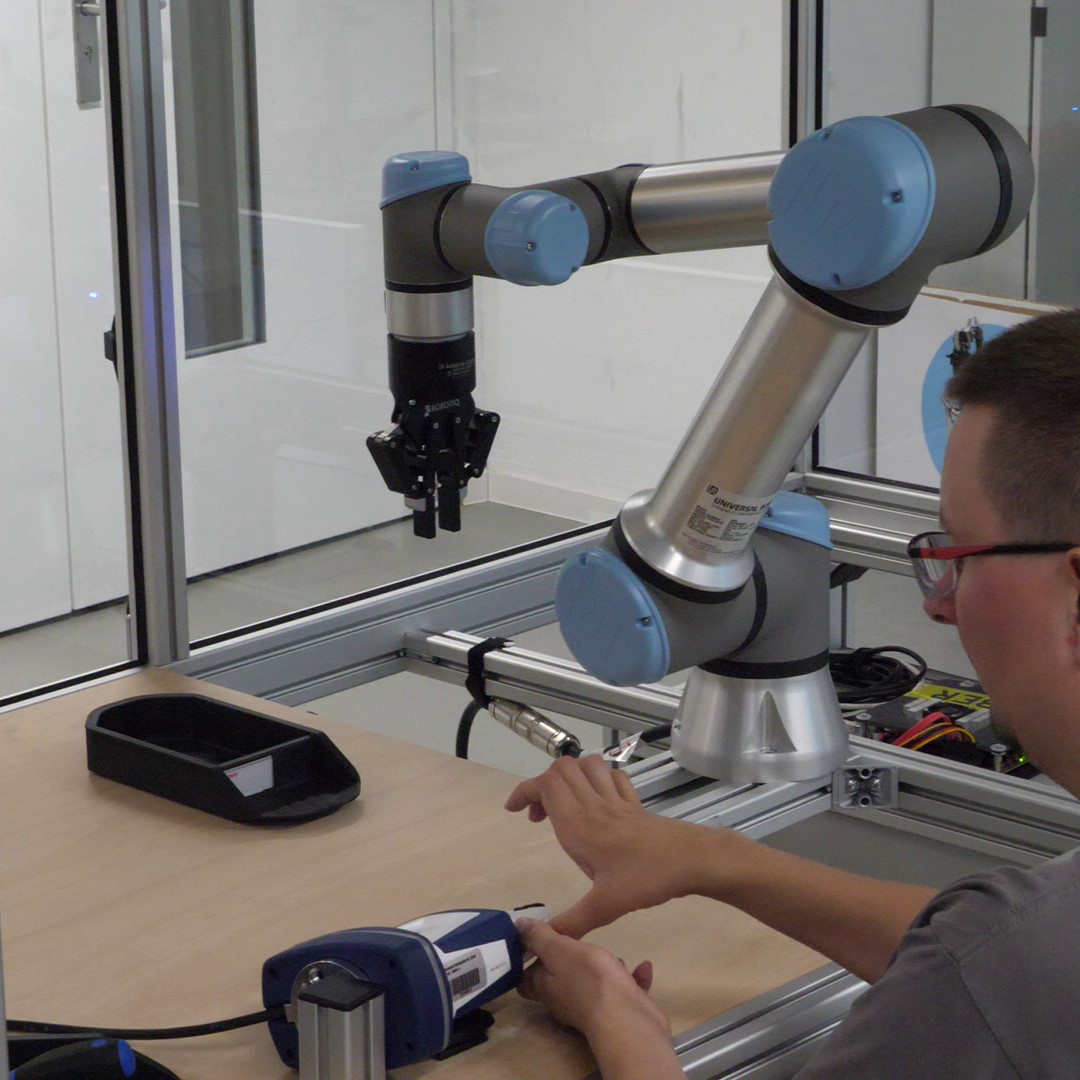}\\
         \vspace{2mm}
         \includegraphics[width=\picwidth]{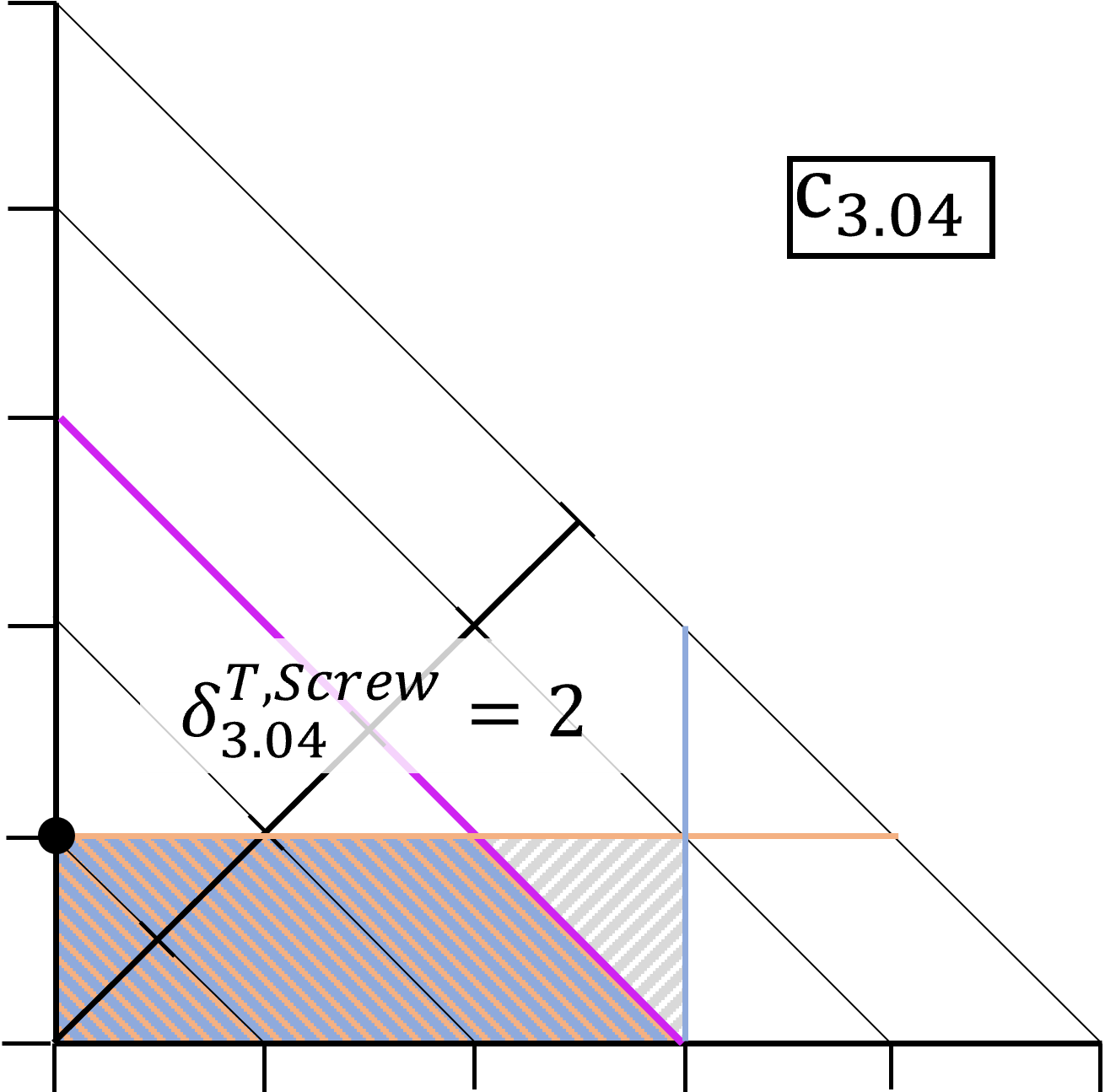}
         \label{sfig:expl_only}
     \end{subfigure}
     \hfill
     \begin{subfigure}[t]{0.245\textwidth}
         \centering
         \includegraphics[width=\toppicwidth]{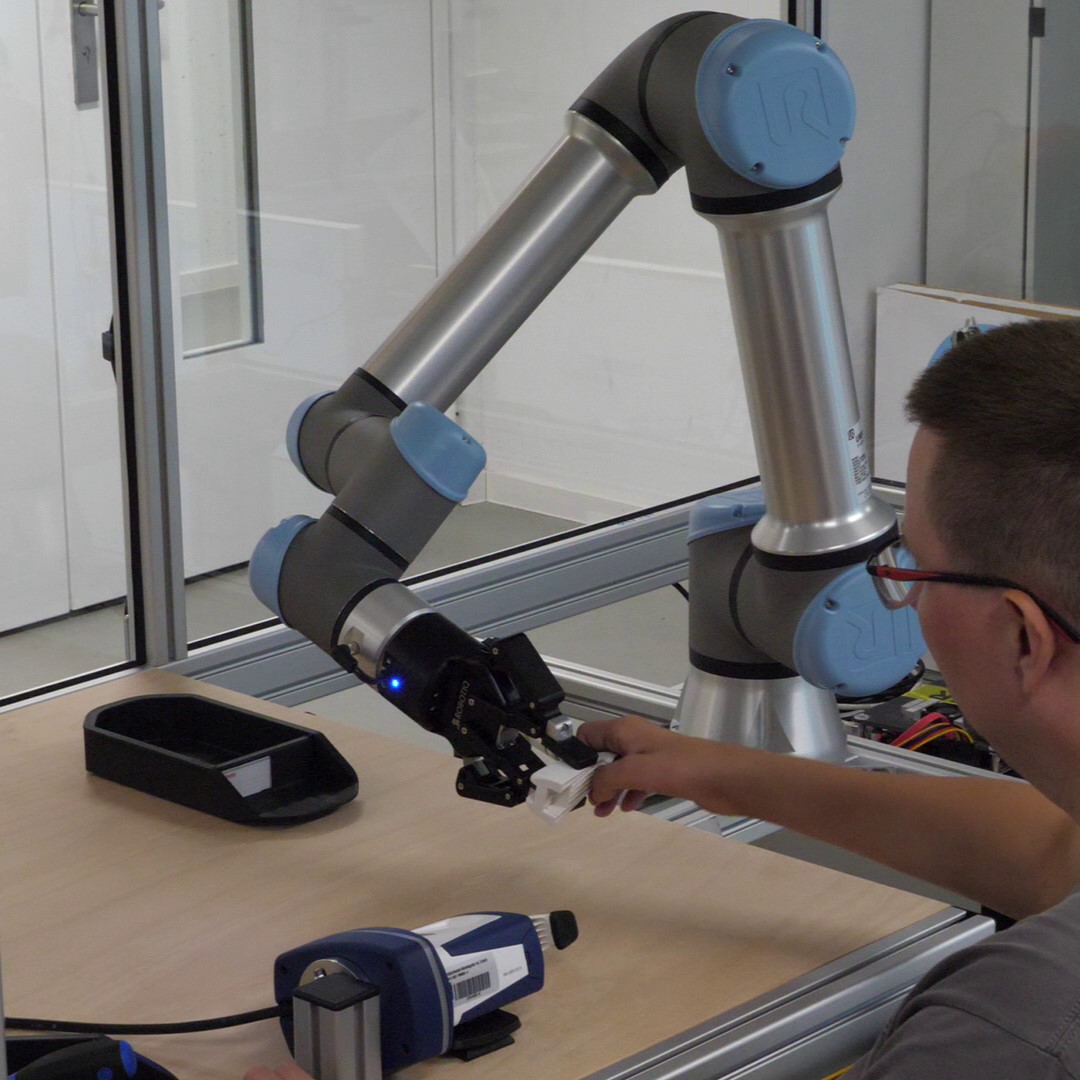}\\
         \vspace{2mm}
         \includegraphics[width=\picwidth]{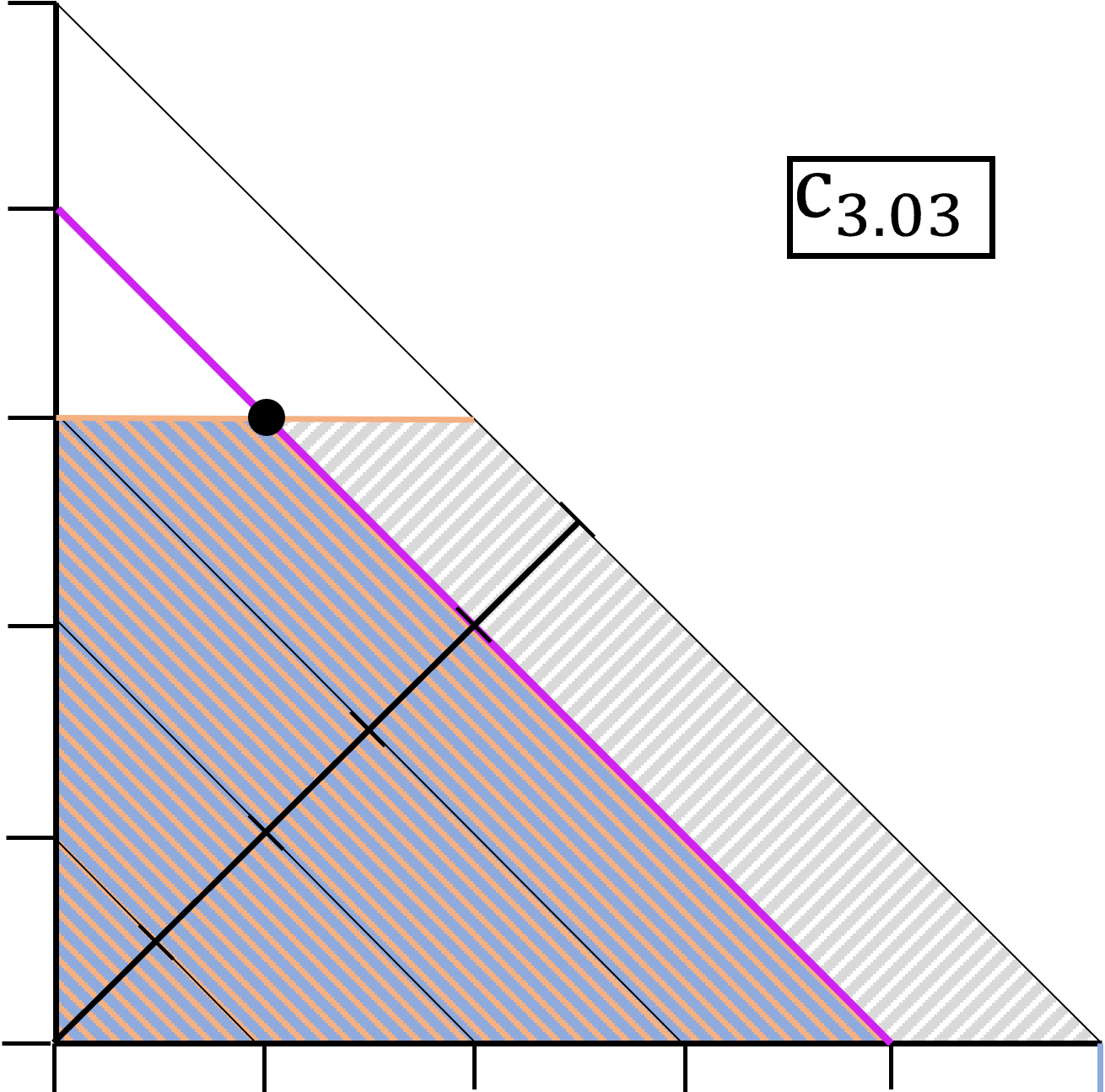}
         \label{sfig:expl_haul_team}
     \end{subfigure}
     \hfill
     \begin{subfigure}[t]{0.245\textwidth}
         \centering
         \includegraphics[width=\toppicwidth]{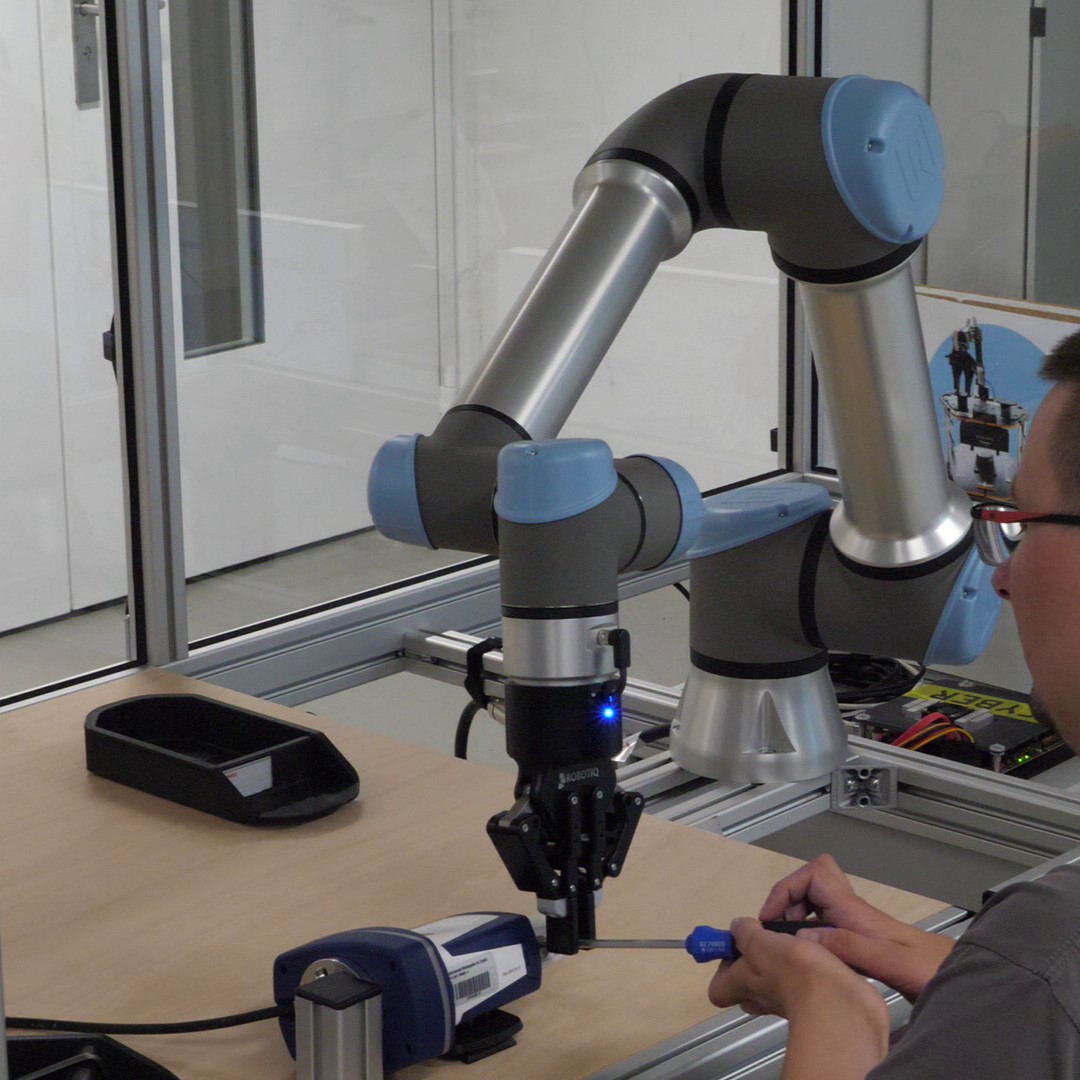}\\
         \vspace{2mm}
         \includegraphics[width=\picwidth]{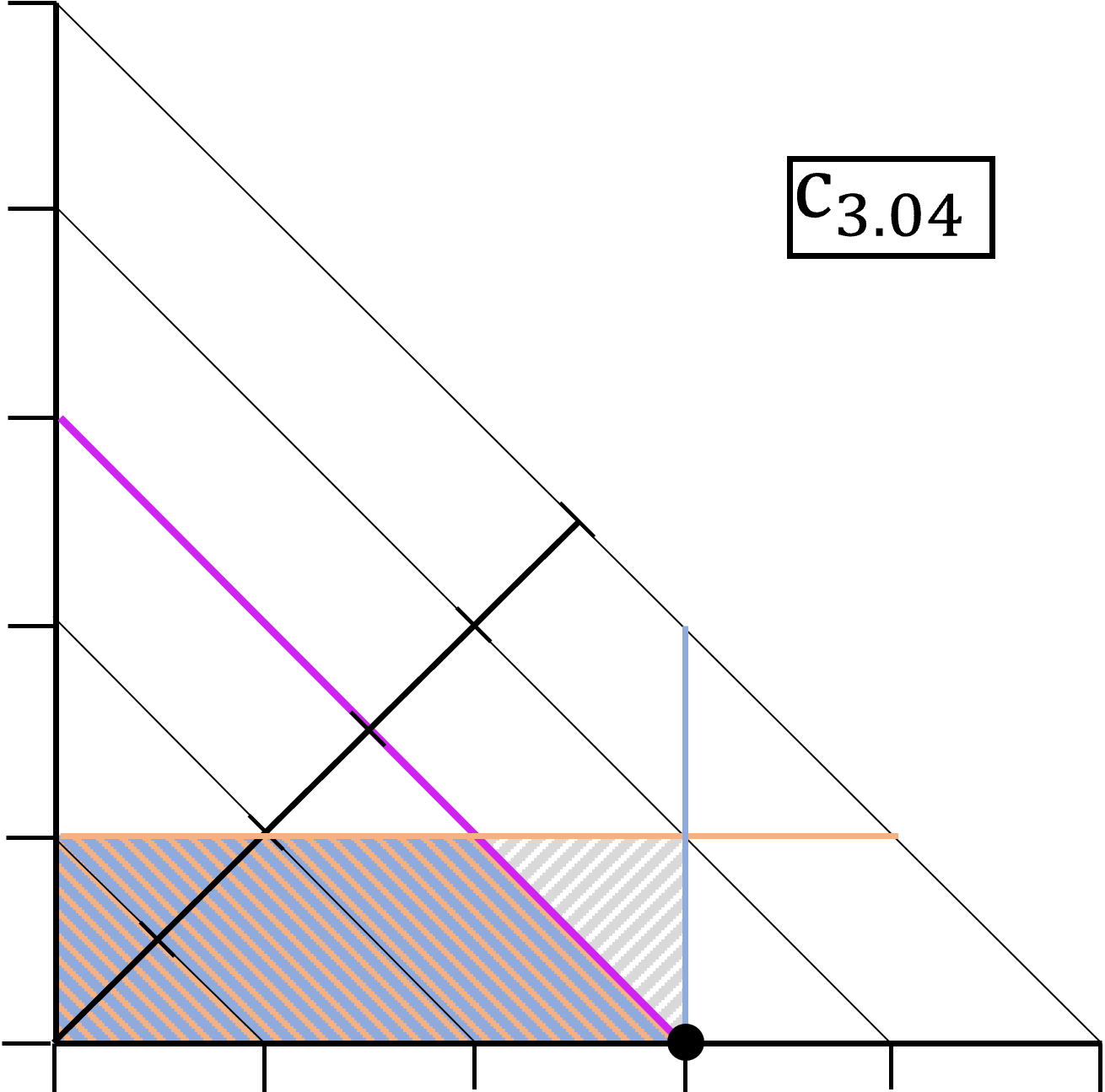}
         \label{sfig:expl_screw_team}
     \end{subfigure}
    \caption{Exploration of manufacturing of a robot gripper, selected capabilities are most influential in the individual task. (1; from left) Participants hauls part. The left arm rests on the table, resulting in bad posture. (2) Participant positions screw. The major disability while pinch gripping influences the twisting motion. (3) Robot supports by handing over the part at pure arm's reach. (4) Robot supports by pre-positioning the screw in the hole. Participant shifts from gripping to screwing.}
    \label{fig:exploration}
\end{figure*}

\section{Exploration}
\label{sec:exploration}
An initial exploration is conducted with a paraplegic participant working with a collaborative robot. The participant is tasked with reaching for and hauling an assembly part as part of a manufacturing task. The part is placed at the limit of the participant's reach (including all body movements). In a second step the participant is tasked with screwing the hauled part onto another assembly. The work process, also including another consequent clipping task, is repeated multiple times with breaks. Afterwards, the participant is interviewed. The exploration is conducted as a fully uncovered Wizard of Oz exploration. Scenes from the exploration and according control distributions are depicted in Figure~\ref{fig:exploration}.

\subsection{Human Only Hauling}
First, the participant is tasked to fulfill the hauling task without artificial assistance. Notably, the participant uses the delta compensation pattern (Section~\ref{sssec:delta_compensation}) to bring the system into a controllable state. As the trunk movement is limited due to the seated posture and his paraplegia, he is far outstretched when reaching for the part, including a support posture with the non-reaching arm. Even though, the posture is non-ergonomic, it is perceived as fulfilling by the participant. This is, as the participant is still relatively young, performative, and self-sufficient as far as possible. Latter is linked with strong \emph{key qualifications} in \emph{IMBA}. Hence, we can deduct that not only physical capabilities may be conjugated, but also physical and mental capabilities (or \emph{key qualifications}). Due to another disability in the fingers, the screwing task is virtually impossible for the participant. Prominently, he lacks fine motor functions to pinch the screw and insert it into the hole.

\subsection{Teamed Tasks}
\label{ssec:teamed_hauling}
Next, the participant is assisted by an Universal Robots UR5e robot in the hauling task. The robot is programmed to grip the part and preposition it at the participant's pure forward reaching limit ($c_{3.03}^{H}=3$), hence, he can reach for the part without applying any trunk movement ($r_{3.02}^{Haul}=0$). The teaming results in a much improved ergonomic posture, compare~\mbox{\cite[p.~8, no.~14]{Gualtieri.2024}}. Unexpectedly, the participant was not in favor of this type of assistance. He dislikes the teamed work process, as the task became too undemanding. This indicates that we were not able to calibrate the correct control distribution. The participant can still move his trunk to a low extent, resulting in a situation with usable resources left in the \emph{trunk movement}. We deduct, that the capabilities shall be used to their full extent as far as possible. The screwing task is adjusted such that the robot pre-positions the screw and the participant is only tasked with operating the screw driver. The participant described this task as fulfilling as he is no longer embarrassed by losing the grip on the screw. He describes the interaction as intuitive. We can deduct that assistance becomes more important with a rising capability delta.
In conclusion, we observed multiple aspects discussed in this paper. Particularly, we were able to show how the theoretical aspects of task fulfillment and over-fulfillment related to capability deltas are mirrored in real-world applications. The application of capability theory and capability deltas makes it better explainable why certain teaming aspects work and others do not. The delta compensation pattern was observed, but more elaborate patterns are missing to tackle all type of teaming challenges (compare Section~\ref{ssec:teamed_hauling}). Further, more research is needed to explore how capability deltas behave in more complex scenarios with higher levels of robot autonomy.

\section{Conclusion}
\label{sec:conclusion}
In this work, we formalized capabilities and their quantification based on occupational analysis with the aim of crafting capability deltas as unified quantifier of the gap which needs to be overcome in teamed task fulfillment. Therefore, we first discussed capability quantification and their dependence on resources. This is used to adapt the arbitration scheme, in which the knowledge of other agents' observed capabilities terms a new base for arbitration. We then matched capabilities with task requirements, which ultimately resulted in the capability deltas. We then discussed the summativity of capabilities and their implication on the control distribution in shared autonomy systems. For choosing suitable control distributions, we introduced the capabilities-requirement diagram, which depicts all priorly discussed aspects. We finally implemented the theory in the delta compensation pattern, which allows to find control distributions over conjugated capabilities. The theory was then subject to an exploration, which supported diverse aspects of the theory, but more work needs to be done to support the findings of this initial study. In future research, we aim to develop a capability estimation method based on the findings of this paper. The theory can be brought to any domain with potential lack in capability (or control), e.g., manufacturing, smart home, elderly care, or (autonomous) vehicles. There is a manifold of users, who may profit from autonomous identification of their individual needs in any form of tasks, and with our theory we bring those systems closer to realization by means of harmonization and unification.

\section*{Acknowledgment}
We like to thank Yi Zhang for her input regarding the \emph{CR} diagram and Burkhard Corves for hosting the exploration.

\bibliographystyle{IEEEtran}
\bibliography{references}

\end{document}